\pdfoutput=1

\documentclass[11pt]{article}
\RequirePackage[color=yellow!20,linecolor=red,textsize=scriptsize]{todonotes}

\usepackage[]{acl}

\usepackage{times}
\usepackage{latexsym}

\usepackage[T1]{fontenc}

\usepackage[utf8]{inputenc}

\usepackage{microtype}

\usepackage{inconsolata}

\usepackage{graphicx}
\usepackage{booktabs, multirow}
\usepackage{subfig}
\usepackage{mathtools}
\usepackage{amsmath}
\usepackage{amsfonts}
\usepackage[inline, shortlabels]{enumitem}
\usepackage{xcolor,colortbl}

\definecolor{myGreen}{rgb}{0.05, 0.5, 0.06}
\definecolor{myRed}{rgb}{0.66, 0.13, 0.24}
\definecolor{myYellow}{rgb}{0.72, 0.53, 0.04}
\definecolor{LightGreen}{rgb}{0.8,1,0.89}
\definecolor{LightCyan}{rgb}{0.88,1,1}

\newcommand{\model}{\textsc{PA3}}
\title{Harmonizing Code-mixed Conversations: Personality-assisted Code-mixed Response Generation in Dialogues}

\author{Shivani Kumar \\
  IIIT Delhi \\
  \texttt{shivaniku@iiitd.ac.in} \\\And
  Tanmoy Chakraborty \\
  IIT Delhi \\
  \texttt{chak.tanmoy.iit@gmail.com} \\}

\begin{document}
\maketitle
\begin{abstract}
Code-mixing, the blending of multiple languages within a single conversation, introduces a distinctive challenge, particularly in the context of response generation. Capturing the intricacies of code-mixing proves to be a formidable task, given the wide-ranging variations influenced by individual speaking styles and cultural backgrounds. In this study, we explore response generation within code-mixed conversations. We introduce a novel approach centered on harnessing the Big Five personality traits acquired in an unsupervised manner from the conversations to bolster the performance of response generation. These inferred personality attributes are seamlessly woven into the fabric of the dialogue context, using a novel fusion mechanism, \model. It uses an effective two-step attention formulation to fuse the dialogue and personality information. This fusion not only enhances the contextual relevance of generated responses but also elevates the overall performance of the model. Our experimental results, grounded in a dataset comprising of multi-party Hindi-English code-mix conversations, highlight the substantial advantages offered by personality-infused models over their conventional counterparts. This is evident in the increase observed in ROUGE and BLUE scores for the response generation task when the identified personality is seamlessly integrated into the dialogue context. Qualitative assessment for personality identification and response generation aligns well with our quantitative results.

\end{abstract}

\section{Introduction}
Conversations\footnote{We use `conversations', `dialogues', and `discourse' interchangeably.} serve as the primary medium for exchanging ideas and cultivating acquaintance among individuals \cite{turnbull2003language}. Remarkably, many people exhibit fluency in multiple languages, seamlessly blending these linguistic resources in their daily communications \cite{tay1989code, tarihoran2022impact}. This phenomenon, characterized by fusing distinct languages to convey meaning, is referred to as {\em code-mixing}. While code-mixing prevails as a widespread linguistic phenomenon \cite{kasper_wagner_2014}, it has not garnered significant attention within the mainstream NLP community, where monolingual text processing has been the predominant focus. Of late, there is a growing recognition of the critical importance of comprehending code-mixed conversations resulting in an increased number of studies investigating diverse aspects of code-mixing in conversations \cite{banerjee-etal-2018-dataset, agarwal-etal-2021-towards, SINGH2022108900, DOWLAGAR2023101449}, such as the identification of humor \cite{khandelwal-etal-2018-humor, bedi2021multi, bukhari2023humor}, emotional expression \cite{ameer2022multi, kumar2023explaining}, and sarcasm \cite{bedi2021multi, kumar-etal-2022-become}. However, the realm of response generation within code-mixed dialogues remains an underexplored frontier \cite{SINGH2022108900}. To this end, we propose tackling the response generation challenge for code-mixed conversations.

\begin{figure}
    \centering
    \includegraphics[width=\columnwidth]{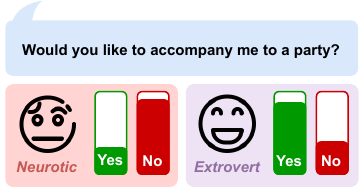}
    \caption{Influence of personality on dialogue responses -- a {\em neurotic} speaker might respond negatively to the posed question, whereas an {\em extrovert} would likely provide a positive reply.}
    \label{fig:personality_example}
    \vspace{-4mm}
\end{figure}

It is crucial to note that while response generation is a vital avenue to explore, it diverges significantly from conventional natural language understanding tasks since a uniform, `one-size-fits-all' model proves inherently inadequate in this context \cite{chen-etal-2020-listeners}. Every individual possesses a unique set of preferences and life experiences, which collectively mould their distinct personalities, subsequently exerting a profound influence on their responses to identical queries \cite{zhang-etal-2018-personalizing}.  Figure \ref{fig:personality_example} illustrates this point. As evident, the response to a seemingly straightforward question, such as {\em ``Would you like to accompany me to a party?''}, can differ based on the listener's prominent personality traits. Interlocutor A, characterized as an \textit{neurotic}, responds distinctively compared to Interlocutor B, who leans more towards being \textit{extrovert}. Appendix \ref{sec:big-5} presents the definition of personality traits, along with examples. 

Personality traits, by their very nature, span a vast spectrum and thus possess the potential for infinite possibilities \cite{alam2014fusion}. Numerous studies have been conducted to quantify these traits \cite{myers1995, https://doi.org/10.1111/j.1758-0854.2008.01007.x, inbook}, with the Big Five personality traits \cite{doi:10.1146/annurev.ps.41.020190.002221} emerging as the prominent framework in this context. This theory distils human personality into five distinctive dimensions: Openness (\textsc{Opn}), Conscientiousness (\textsc{Con}), Extraversion (\textsc{Ext}), Agreeableness (\textsc{Agr}), and Neuroticism (\textsc{Neu}), in which each dimension encapsulates a pivotal facet of an individual's character. For instance, elevated levels of openness may signify a predisposition towards imagination. Here, we adopt this widely accepted model as the foundation for characterizing a speaker's personality. Our central hypothesis contends that incorporating personality indicators within the response generation process plays a pivotal role in generating contextually appropriate responses to given queries. Given the intricate and non-generalizable nature of manually annotating personality traits, we propose an unsupervised learning approach to acquire these traits, which, in turn, enhances response generation capabilities. In a nutshell, our contributions are four-fold:
\begin{enumerate}[leftmargin=*,noitemsep,topsep=0pt]
    \item We explore the task of \textbf{code-mixed response generation}.
    \item We propose an \textbf{unsupervised mechanism to identify speakers' personality traits} and leverage them for better response generation.
    \item We propose \textbf{a novel method, \model}\footnote{\textsc{P}ersonality-\textsc{A}ware \textsc{A}xial \textsc{A}ttention}, which combines the identified traits with dialogue context to generate responses.
    \item Our \textbf{quantitative and qualitative analyses} show the benefits of including personality traits in code-mixed response generation.
\end{enumerate}

\section{Related Works}
\paragraph{Conversation and Code-mixing.} Dialogues represent a well-established domain in NLP, having undergone extensive exploration \cite{10.1145/3166054.3166058, kumar2023dialogue}. However, the bulk of this research has predominantly revolved around monolingual text, despite the enduring prevalence of code-mixing, a timeworn linguistic phenomenon \cite{tay1989code}. Consequently, recent years have witnessed a surge in studies dedicated to unravelling the intricacies of code-mixing within dialogues \cite{ahn2020code}. These investigations have honed in on exploring various nuances of code-mixed dialogues, delving into attributes such as intents \cite{Liu_Winata_Lin_Xu_Fung_2020, FIRDAUS2023299}, the presence of hate speech \cite{modha2021overview, madhu2023detecting}, humor \cite{khandelwal-etal-2018-humor, bedi2021multi}, and sarcasm \cite{bedi2021multi,kumar-etal-2022-become}. Yet, the landscape for the generative dimension of code-mixing remains relatively uncharted, with limited concerted efforts in this direction.

\textbf{Response Generation.} For dialogue agents, it is of paramount importance to keep the conversation engaging \cite{alexabot}. Consequently, generating apt responses becomes a primary field of research in terms of dialogue analysis. Many studies have been conducted to generate the right responses for monolingual English dialogues \cite{spring2019empathic, fan2020survey, 10.1145/3554727}. However, response generation in the code-mixed setting remains a comparatively unexplored topic with only a handful of existing studies \cite{agarwal-etal-2021-towards, SINGH2022108900}. \citet{10.1145/3392846} illustrated that multilingual speakers prefer chatbots that can code-mix, thus making code-mixed response generation crucial.

\textbf{Big Five Personality Traits.} In pursuit of a deeper understanding of the user's personality, a range of studies have delved into the realm of the Big Five personality  \cite{costa1992normal,costa2008revised}. Numerous studies endeavored to categorize individuals into one of these personality archetypes based on their salient attributes \cite{mairesse2007using, golbeck2011predicting, kosinski2013private, schwartz2013personality}. A few studies have also attempted to use different personality theories other than the Big Five personality traits such as  MBTI \cite{myers1995,celli2018big}.

\textbf{Personality-assisted Response Generation.} The significance of personalization in enhancing the efficacy of dialogue systems is widely acknowledged  \cite{lucas2009managing, joshi2017personalization, weston2018retrieve, dinan2018wizard, roller2020recipes, chen2020listener}. While earlier studies primarily concentrated on the utilization of user profiles to tailor goal-oriented dialogue systems \cite{lucas2009managing,joshi2017personalization}, recent investigations have shifted their focus towards chit-chat settings \cite{li2016persona, zhang2018personalizing, weston2018retrieve,roller2020recipes,dinan2018wizard}. However, all of these studies deal with monolingual data. Consequently, we explore personality-assisted response generation in a code-mixed setting.

\section{Problem Definition}
The complete problem definition can be divided into two phases as follows:

\textbf{Phase 1: Speaker Personality Detection.} Given the contextual utterances $(s_1, u_1), (s_2, u_2), \dots, (s_{n-1}, u_{n-1})$ such that utterance $u_i$ is uttered by speaker $s_j$, we aim to generate personality $p_n$ for speaker $s_n$.
A classification model selects $p_n$, such that $p_n \in P$ and $P = $\{\textsc{Opn}, \textsc{Con}, \textsc{Ext}, \textsc{Agr}, \textsc{Neu}\}, and maps the selected trait class into a templatic personality defining the speaker (c.f. Table \ref{tab:personality_template}). We append this definition with the input and move on to phase 2.

\textbf{Phase 2: Response generation.} Along with the contextual utterances, the input also contains the personality trait for the subsequent speaker, such that the input becomes $\{(s_1, u_1), (s_2, u_2), \dots, (s_{n-1}, u_{n-1}), p_n\}$. Response generation aims to generate utterance $u_n$ by speaker $s_n$ based on the detected personality $p_n$.

\section{Dataset}
\begin{figure}[t]
\centering
    \subfloat[Statistics of MaSaC.]{
        \resizebox{\columnwidth}{!}{%
        \begin{tabular}{l|c|c|c|cc|ccc}\toprule
            \multirow{2}{*}{\textbf{Set}} &\multirow{2}{*}{\textbf{\#Dlgs}} &\multirow{2}{*}{\textbf{\#Utts}} &\multirow{2}{*}{\textbf{Avg sp/dlg}} &\multicolumn{2}{c}{\textbf{Utt len}} &\multicolumn{2}{|c}{\textbf{Vocab len}} \\\cmidrule{5-8}
            
            & & & &\textbf{Avg} &\textbf{Max} &\textbf{English} &\textbf{Hindi} \\\midrule
            
            \textbf{Train} &8506 &8506 &3.60 &10.82 &113 &\multirow{4}{*}{3157} &\multirow{4}{*}{14803} \\
            
            \textbf{Val} &45 &1354 &4.13 &10.12 &218 & & \\
            
            \textbf{Test} &56 &1580 &4.32 &10.61 &84 & & \\
            
            \cmidrule{1-6}
            
            \textbf{Total} &8607 &11440 &12.05 &31.55 &415 & & \\
            
            \bottomrule
        \end{tabular}
        \label{tab:data_stats}
    }}\\\vspace{-2mm}
    \resizebox{0.47\columnwidth}{!}{%
    \subfloat[Speaker distribution in the MaSaC dataset.]{
        \includegraphics[width=0.47\columnwidth]{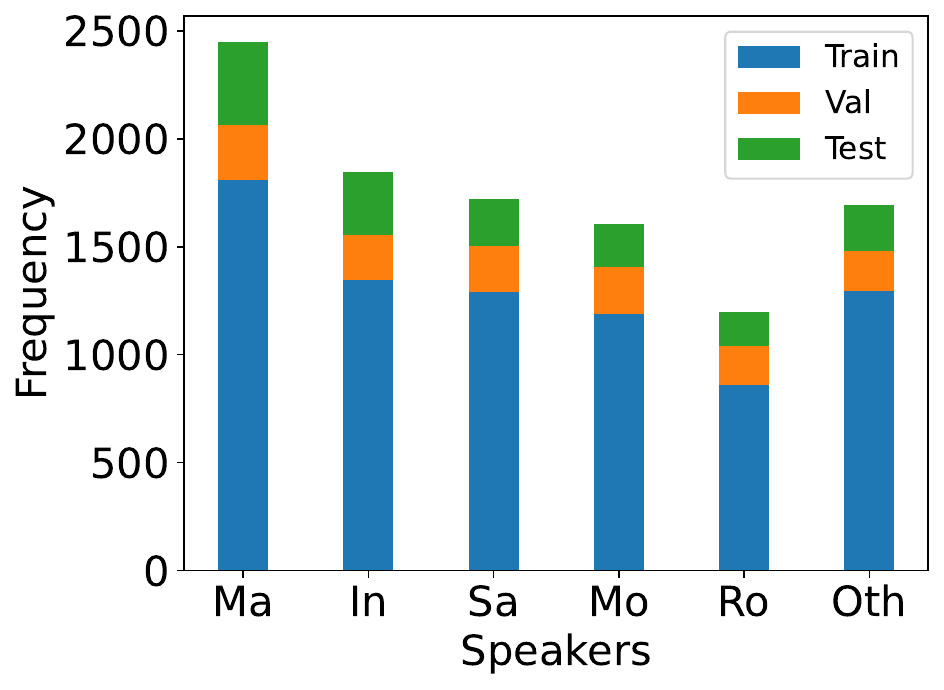}
        \label{fig:data_sp_dist}
    }}
    \resizebox{0.47\columnwidth}{!}{%
    \subfloat[Number of speakers other than the five primary speakers.]{
        \includegraphics[width=0.47\columnwidth]{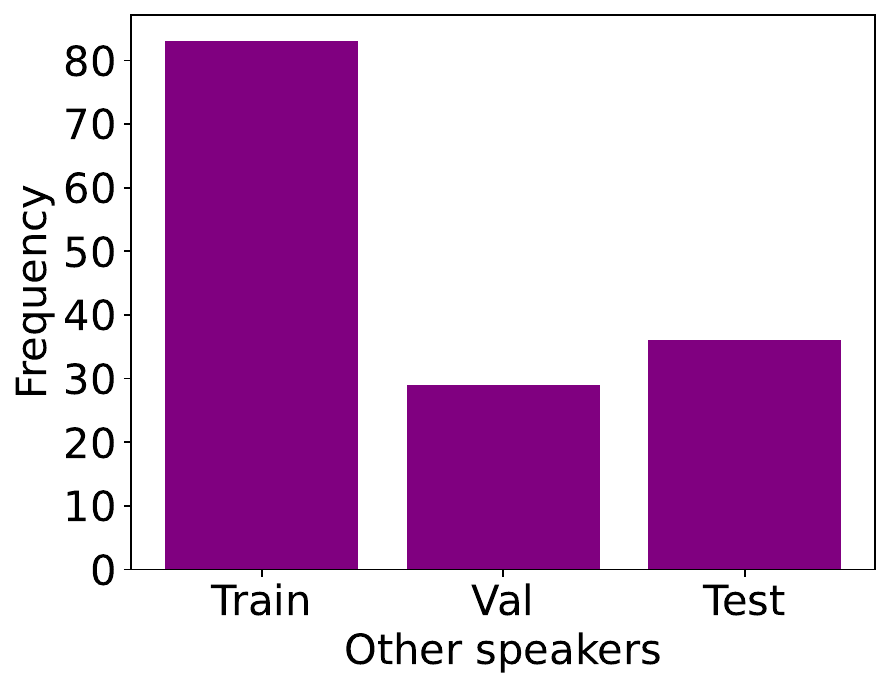}
        \label{fig:data_sp_other}
    }}\vspace{-1mm}
\caption{Dataset description of MaSaC (Abbreviation: Dlgs: Dialogues, Utts: Utterances, sp: speakers, Ma: Maya, In: Indravardhan, Sa: Sahil, Mo: Monisha, Ro: Rosesh, Oth: Others).}
\label{fig:data-stats}
\vspace{-4mm}
\end{figure}
Datasets for code-mixed conversations are inadequate, especially for multi-turn, multi-party conversations. In this study, we consider the MaSaC dataset \cite{bedi2021multi}, containing Hindi-English code-mixed discourse of multi-turn and multi-party nature from an Indian TV series\footnote{\url{https://www.imdb.com/title/tt1518542/}}. The dataset was originally curated to perform the task of sarcasm and humour detection since it contains conversations similar to daily discourse among peers. Consequently, we extract the conversations from this dataset and construct our response generation instances. We highlight the critical statistics of the dataset in Table \ref{tab:data_stats}. The speaker distribution in Figure \ref{fig:data_sp_dist} and Figure \ref{fig:data_sp_other} shows that there are five primary speakers in the dataset, each with varying personalities (c.f. Table \ref{tab:pred_personality}). Thus, aiding response generation with speaker personalities can improve its performance.

\section{Proposed Methodology}
\begin{figure*}[ht]
    \centering
    \includegraphics[width=\textwidth]{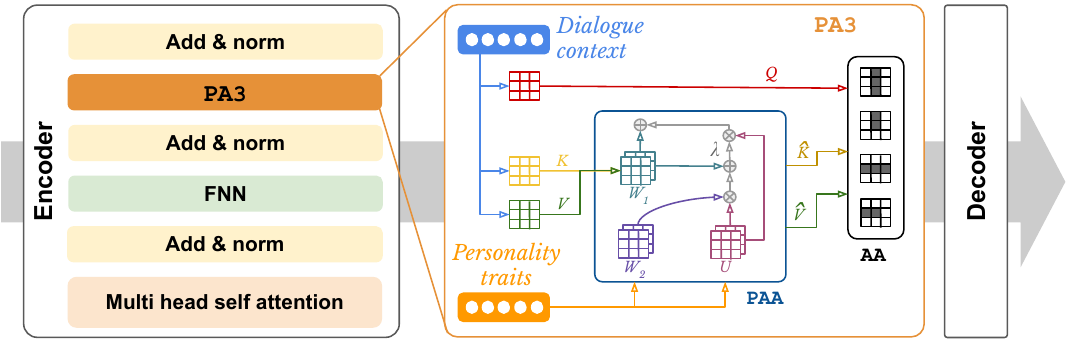}
    \caption{Model architecture to fuse personality values with dialogue context. The \model\ module can be injected into any encoder-decoder architecture, and it takes as inputs the dialogue representation along with the personality trait definition representation. First, context-aware attention is used to learn personality-infused key and value pairs and axial attention is then used to combine query, key, and value vectors into one final representation.}
    \label{fig:model}
    \vspace{-3mm}
\end{figure*}

In this section, we discuss our proposed methodology, with the foremost objective being the effective identification of personality attributes from the dialogue context. To achieve this, we propose an unsupervised technique that leverages response generation performance to improve personality identification. Subsequently, we fuse the personality attributes into the dialogue context to generate responses influenced by individual traits. We propose the incorporation of an intermediary module within the core encoder. This module leverages a straightforward yet effective two-step attention mechanism, facilitating the fusion of personality attributes with the representation of the dialogue.
Broadly, we employ context-aware attention \cite{Yang_Li_Wong_Chao_Wang_Tu_2019}, which is instrumental in infusing personality characteristics into the key and value vectors of the dialogue. Subsequently, we employ Axial attention \cite{ho2020axial} to yield a refined, conclusive representation, which ultimately feeds into the decoder. 
Figure \ref{fig:model} provides a schematic diagram of our model. In the following subsections, we offer a comprehensive overview of individual modules.

\begin{figure}[t]
    \centering
    \includegraphics[width=\columnwidth]{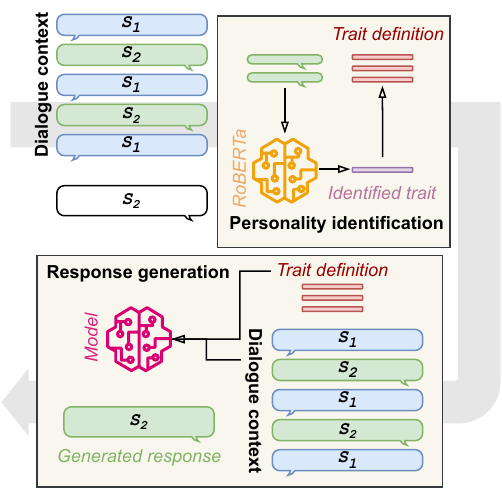}
    \caption{Outline of learning personality traits using the \textit{`pseudo'} task of response generation.}
    \label{fig:word2vecTraits}
    \vspace{-4mm}
\end{figure}

\subsection{Personality Identification}
In this section, we describe our methodology for discerning the personality traits of each speaker and subsequently mapping them to their corresponding trait definitions. Although multiple theories quantify a speaker's personality traits \cite{myers1995, https://doi.org/10.1111/j.1758-0854.2008.01007.x, inbook}, existing NLP applications widely use the Big Five Personality theory \cite{doi:10.1146/annurev.ps.41.020190.002221}. Consequently, we select this model for our study, encompassing five distinct personality dimensions as shown in Table \ref{tab:personality_template}, where one of these dimensions is presumed to be dominant. To find the most suitable personality trait for a speaker in a dialogue, we employ an approach similar to  Word2Vec  \cite{mikolov2013efficient}, where a \textit{`pseudo'} task is implemented to facilitate the acquisition of word embeddings. In the context of personality identification, our \textit{`pseudo'} task takes the form of response generation, where we seek to enhance the generated response based on the intermediary step of personality identification. Figure \ref{fig:word2vecTraits} gives an overview of our mechanism for personality identification. We employ RoBERTa base \cite{liu2020roberta} to classify personalities attributed to the target speaker, using the input dialogue as the primary data source. Once the personality is identified, it is subsequently linked to its templatic definition --- a descriptive representation of the speaker's character, as outlined in Table \ref{tab:personality_template}. This personality definition is presented alongside the input dialogue to an encoder for further steps in the proposed pipeline.

\begin{table}[t]\centering
\resizebox{\columnwidth}{!}{
    \begin{tabular}{l|p{20em}r}\toprule
    \textbf{Trait} &\textbf{Templatic Definition} \\\toprule
    
    Openness &The speaker has high openness trait. They embrace new ideas, are curious about the world, and are often drawn to creative and unconventional pursuits. \\ \midrule
    
    Conscientious &The speaker has conscientiousness trait. They are reliable, organized, and detail-oriented, demonstrating a strong work ethic and a commitment to achieving their goals. \\ \midrule
    
    Extraversion &The speaker has extraversion trait. They thrive in social settings, energized by interactions with others, and enjoy being at the center of activities. \\ \midrule
    
    Agreebleness &The speaker has agreeableness trait. They prioritize cooperation, are empathetic, and often go out of their way to maintain harmonious relationships and help others. \\ \midrule
    
    Neuroticism &The speaker has high neuroticism trait. They have a greater tendency for emotional instability, anxiety, and a propensity to experience negative emotions such as fear, sadness, and anger. \\
    \bottomrule
    \end{tabular}
}
\caption{Personality traits in the Big Five personality model along with their templatic definitions.}
\label{tab:personality_template}
\vspace{-4mm}
\end{table}

\subsection{Personality-Aware Attention (\textsc{PAA})}
With the personality definition and the input dialogue at our disposal, our next step is to seamlessly integrate the personality information with the dialogue information to craft a suitable response. Conventional attention-based fusion mechanisms often facilitate a direct interplay between the input representations, in which one representation functions as the query while the others assume the roles of key and value. However, as each representation captures distinct attributes, straightforward fusion may not preserve the optimal contextual information and could introduce significant noise into the final representations. Consequently, we introduce personality-aware attention (\textsc{PAA}) fusion employing context-aware attention \cite{Yang_Li_Wong_Chao_Wang_Tu_2019}. Our method entails the initial generation of personality-conditioned key and value vectors, followed by applying axial attention \cite{ho2020axial} to obtain the final fused values. We explain the process in detail below.

For an encoder model, we have the intermediate representation $H$ at a specific layer to compute the query, key, and value vectors denoted as $Q$, $K$, and $V$ respectively, in $\mathbb{R}^{n \times d}$ as outlined in Equation~\ref{eq:qkv}. $W_Q, W_K, and W_V$ are model parameters each with dimensions of $\mathbb{R}^{d \times d}$. In this context, $n$ signifies the maximum sequence length of the text, while $d$ represents the dimensionality of the dialogue vector.
\begin{eqnarray}
    \begin{bmatrix}
    Q  K  V
    \end{bmatrix} = H \begin{bmatrix}
    W_Q  W_K  W_V
    \end{bmatrix}
    \label{eq:qkv}
\end{eqnarray}

The vector $P$ in $\mathbb{R}^{n \times d_p}$, the encoded personality vector is used to create personality-influenced key and value vectors, $\hat K$ and $\hat V$ respectively, based on the method outlined by \citet{Yang_Li_Wong_Chao_Wang_Tu_2019}. For balancing of information from the personality source and information retention from the dialogue, we train a matrix $\lambda$ in $\mathbb{R}^{n \times 1}$ (Equation \ref{eq:lambda}). $U_k$ and $U_v$ in $\mathbb{R}^{d_p \times d}$ are matrices that can be learned.
\begin{eqnarray}
    \begin{bmatrix}
    \hat K \\ \hat V
    \end{bmatrix} = (1 - \begin{bmatrix}
    \lambda_k \\ \lambda_v
    \end{bmatrix})\begin{bmatrix}
    K \\ V
    \end{bmatrix} + \begin{bmatrix}
    \lambda_k \\ \lambda_v
    \end{bmatrix}(P\begin{bmatrix}
    U_k \\ U_v
    \end{bmatrix})
\end{eqnarray}

Rather than setting $\lambda_k$ and $\lambda_v$ as hyperparameters, we allow the model to autonomously determine their values through a gating mechanism, as defined in Equation~\ref{eq:lambda}. Additionally, the matrices $W_{k_1}, W_{k_2}, W_{v_1},$ and $W_{v_2}$, each with dimensions $\mathbb{R}^{d \times 1}$, are trained in conjunction with the model.

\vspace{-1em}
\begin{eqnarray}
    \begin{bmatrix}
    \lambda_k \\ \lambda_v
    \end{bmatrix} = \sigma(\begin{bmatrix}
    K \\ V
    \end{bmatrix} \begin{bmatrix}
    W_{k_1} \\ W_{v_1}
    \end{bmatrix} + P\begin{bmatrix}
    U_k \\ U_v
    \end{bmatrix} \begin{bmatrix}
    W_{k_2} \\ W_{v_2}  
    \end{bmatrix})
    \label{eq:lambda}
\end{eqnarray}

Once we obtain the personality-infused key and value vectors, we use the Axial attention mechanism as described below.

\subsection{Axial Attention}
Axial attention \cite{ho2020axial} finds its primary application in computer vision, where its utility extends to managing multidimensional tensors. The fundamental aim is to approach each axis independently, thereby comprehensively exploring relationships between the various dimensions. The proposed approach preserves the original shape of the multidimensional tensor, performing either masked or unmasked attention along a single axis at any given time. This specific operation, referred to as axial attention and denoted as Attention$_k$($x$), is responsible for directing attention over axis $k$ within the tensor $x$. In doing so, it blends information across axis $k$ while maintaining the independence of information along the remaining axes. Implementing axial attention for a given axis $k$ involves a series of steps, such as transposing all axes except $k$ to the batch axis, invoking standard attention as a subroutine, and reverting the transpose operation. Within our network architecture, we leverage two axial attention layers, culminating in the derivation of the ultimate dialogue representation denoted as $\hat H$, signifying the personality-infused representation of the dialogue, which is then passed on to the next encoder/decoder layer. For our input two dimensional arrays of $\hat K$, $\hat V$, and $Q$:
\begin{equation}
    \hat H = \text{Attention}_k(\hat K, \hat V, Q)
\end{equation}

\section{Experiments and Results}
\paragraph{Evaluation Metrics.}  Given the absence of ground-truth labels for evaluating personality detection, we resort to a manual assessment process, meticulously scrutinizing the outputs for the primary speakers to derive meaningful insights into the system's performance in this regard. To assess the response generation proficiency, we employ established evaluation metrics, specifically ROUGE \cite{lin-2004-rouge} and BLEU \cite{papineni2002bleu} scores. These metrics are adept at quantifying the syntactic competence of the system in question. Additionally, we incorporate BERTScore \citep{zhang2019bertscore}, which serves to gauge the semantic aptitude of the system, and human evaluation provides a more comprehensive evaluation.

In this section, we present a comprehensive overview of the quantitative and qualitative results achieved by personality identification and response generation. Additionally, we offer a closer look at our ablation results, shedding light on the significance of each submodule within our proposed architectural framework for response generation. Further, human evaluation highlights the pros and cons of the generated responses and personalities.

\subsection{Personality Identification}
\begin{figure}[t]
    \centering
    \includegraphics[width=0.8\columnwidth]{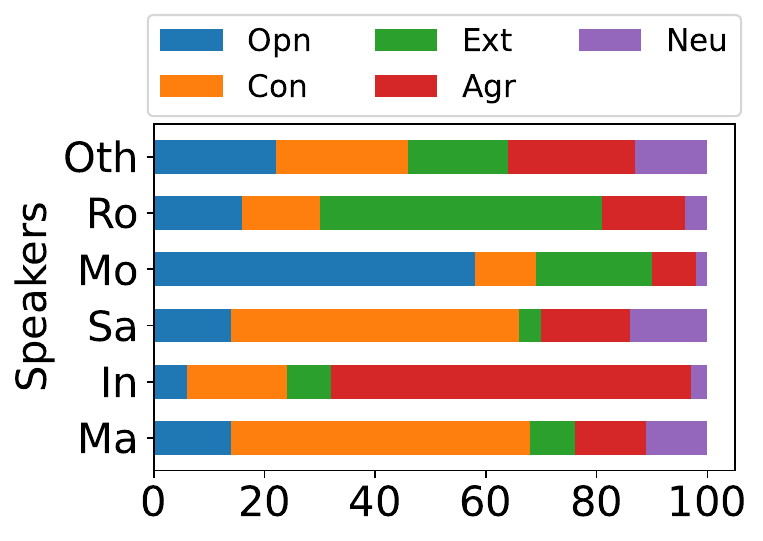}
    \caption{Distribution of the predicted personality traits assigned to different speakers (Abbr - Ma: Maya, In: Indravardhan, Sa: Sahil, Mo: Monisha, Ro: Rosesh, Oth: Others).}
    \label{fig:personality_results}
    \vspace{-4mm}
\end{figure}

As shown in Figure \ref{fig:word2vecTraits}, our initial step predicts the most suitable personality from the Big Five personality traits for the target speaker. To gauge the efficacy of our predicted personalities, we focus on the five primary speakers featured in the MaSaC dataset. Figure \ref{fig:data_sp_dist} shows the distribution of the speakers where it can be observed that the speakers --- Maya, Indravardhan, Sahil, Monisha, and Rosesh, are the most frequently occurring speakers. We perform a manual evaluation of the personality predictions. Using information from Wikipedia\footnote{\url{https://en.wikipedia.org/wiki/Sarabhai_vs_Sarabhai}}, we procure character descriptions for each of the five prominent speakers (c.f. Appendix \ref{sec:speaker_char}) which were given to five expert annotators. The annotators then categorize each speaker within the Big Five personality framework. More information can be found in Appendix \ref{sec:pers_id}. This annotator-driven classification enables the construction of a definitive ground-truth for evaluation encompassing the five speakers, each associated with an assigned personality trait value as shown in Table \ref{tab:pred_personality}. We compare the obtained ground-truth personalities with the ones predicted by the RoBERTa model, an outcome of the \textit{`pseudo'} task centred around response generation. The ensuing distribution of these predictions is laid out for scrutiny in both Table \ref{tab:pred_personality} and Figure \ref{fig:personality_results}. We can see that the personalities found most suitable by the human annotators are the ones preferred by the RoBERTa model, too, validating the performance of our system.

\begin{table}[t]\centering
\resizebox{\columnwidth}{!}{
    \begin{tabular}{l|c|rrrrrr}\toprule
    \textbf{Sp} &\textbf{GT} &\textbf{\textsc{Opn}} &\textbf{\textsc{Con}} &\textbf{\textsc{Ext}} &\textbf{\textsc{Ext}} &\textbf{\textsc{Neu}} \\\midrule
    \textbf{Ma} & \textsc{Con} &14\% &\cellcolor{blue!20}\textbf{54\%} &8\% &13\% &11\% \\
    \textbf{In} & \textsc{Agr} &6\% &18\% &8\% &\cellcolor{blue!20}\textbf{65\%} &3\% \\
    \textbf{Sa} & \textsc{Con} &14\% &\cellcolor{blue!20}\textbf{52\%} &4\% &16\% &14\% \\
    \textbf{Mo} & \textsc{Opn} &\cellcolor{blue!20}\textbf{58\%} &11\% &21\% &8\% &2\% \\
    \textbf{Ro} & \textsc{Ext} &16\% &14\% &\cellcolor{blue!20}\textbf{51\%} &15\% &4\% \\
    \bottomrule
    \end{tabular}
}
\caption{Percentage of times a personality trait is assigned to a speaker. (Abbr - Sp: Speakers, GT: Ground Truth, Ma: Maya, In: Indravardhan, Sa: Sahil, Mo: Monisha, Ro: Rosesh, Oth: Others)}
\label{tab:pred_personality}
\vspace{-4mm}
\end{table}
\begin{table*}[t]\centering
\resizebox{\textwidth}{!}{
    \begin{tabular}{l|l|rrrrrrrrr}\toprule
    \multicolumn{2}{c|}{\textbf{Model}} &\textbf{R1} &\textbf{R2} &\textbf{RL} &\textbf{B1} &\textbf{B2} &\textbf{B3} &\textbf{B4} &\textbf{BS} \\\midrule
    
    \multirow{6}{*}{\rotatebox{90}{\textbf{w/o personality}}} &\textbf{RNN} &8.17 &0.02 &8.09 &5.11 &0.01 &0.11 &0 &54.16 \\
    
    &\textbf{PGN} &7.06 &0 &7.01 &4.31 &0 &0.08 &0 &53.12 \\
    
    &\textbf{Transformers} &10.64 &0.83 &10.35 &7.22 &0.92 &0.13 &0.01 &58.94 \\

    &\textbf{mBART} &11.36 &1.23 &10.9 &7.91 &1.01 &0.21 &0 &61.02 \\
    
    &\textbf{T5} &11.87 &1.01 &11.43 &8.41 &1.02 &0.17 &0.02 &61.98 \\
    
    &\textbf{BART} &12.94 &1.66 &12.34 &9.66 &1.64 &0.43 &0.07 &63.12 \\ \midrule
    
    \multirow{14}{*}{\rotatebox{90}{\textbf{w personality}}} &\textbf{RNN$_{\text{\model}}$} &9.96 {\color{myGreen}($\uparrow$1.79)} &0.08 {\color{myGreen}($\uparrow$0.06)} &10.71 {\color{myGreen}($\uparrow$2.62)} &6.87 {\color{myGreen}($\uparrow$1.76)} &1.04 {\color{myGreen}($\uparrow$1.03)} &0.43 {\color{myGreen}($\uparrow$0.32)} &0.22 {\color{myGreen}($\uparrow$0.22)} &56.24 {\color{myGreen}($\uparrow$2.08)} \\
    
    &\textbf{PGN$_{\text{\model}}$} &8.45 {\color{myGreen}($\uparrow$1.39)} &1.11 {\color{myGreen}($\uparrow$1.11)} &9.41 {\color{myGreen}($\uparrow$2.40)} &5.95 {\color{myGreen}($\uparrow$1.64)} &1.03 {\color{myGreen}($\uparrow$1.03)} &0.37 {\color{myGreen}($\uparrow$0.29)} &0.21 {\color{myGreen}($\uparrow$0.21)} &55.87 {\color{myGreen}($\uparrow$2.75)} \\
    
    &\textbf{Transformers$_{\text{\model}}$} &12.76 {\color{myGreen}($\uparrow$2.12)} &1.75 {\color{myGreen}($\uparrow$0.92)} &12.14 {\color{myGreen}($\uparrow$1.79)} &8.46 {\color{myGreen}($\uparrow$1.24)} &2.02 {\color{myGreen}($\uparrow$1.10)} &0.45 {\color{myGreen}($\uparrow$0.32)} &0.24 {\color{myGreen}($\uparrow$0.23)} &61.06 {\color{myGreen}($\uparrow$2.12)} \\
    
    &\textbf{mBART$_{\text{\model}}$} &13.43 {\color{myGreen}($\uparrow$2.07)} &2.36 {\color{myGreen}($\uparrow$1.13)} &12.15 {\color{myGreen}($\uparrow$1.25)} &8.89 {\color{myGreen}($\uparrow$0.98)} &\textbf{2.61} {\color{myGreen}($\uparrow$1.60)} &0.56 {\color{myGreen}($\uparrow$0.35)} &0.18 {\color{myGreen}($\uparrow$0.18)} &63.42 {\color{myGreen}($\uparrow$2.40)} \\\cmidrule{2-10}

    &\textbf{T5$_{\text{SC}}$} &12.02 {\color{myGreen}($\uparrow$0.15)} &1.51 {\color{myGreen}($\uparrow$0.50)} &11.98 {\color{myGreen}($\uparrow$0.55)} &8.52 {\color{myGreen}($\uparrow$0.11)} &1.51 {\color{myGreen}($\uparrow$0.49)} &0.39 {\color{myGreen}($\uparrow$0.22)} &0.11 {\color{myGreen}($\uparrow$0.09)} &62.05 {\color{myGreen}($\uparrow$0.07)} \\
    
    &\textbf{T5$_{\text{DPA}}$} &12.04 {\color{myGreen}($\uparrow$0.17)} &1.56 {\color{myGreen}($\uparrow$0.55)} &12.01 {\color{myGreen}($\uparrow$0.58)} &8.58 {\color{myGreen}($\uparrow$0.17)} &1.58 {\color{myGreen}($\uparrow$0.56)} &0.41 {\color{myGreen}($\uparrow$0.24)} &0.14 {\color{myGreen}($\uparrow$0.12)} &62.35 {\color{myGreen}($\uparrow$0.37)} \\
    
    &\textbf{T5$_{\text{\model}-\text{Axial}}$} &12.79 {\color{myGreen}($\uparrow$0.92)} &1.64 {\color{myGreen}($\uparrow$0.63)} &12.53 {\color{myGreen}($\uparrow$1.10)} &9.04 {\color{myGreen}($\uparrow$0.63)} &1.96 {\color{myGreen}($\uparrow$0.94)} &0.46 {\color{myGreen}($\uparrow$0.29)} &0.18 {\color{myGreen}($\uparrow$0.16)} &62.99 {\color{myGreen}($\uparrow$1.01)} \\
    
    &\textbf{T5$_{\text{OT}}$} &13.48 {\color{myGreen}($\uparrow$1.61)} &1.97 {\color{myGreen}($\uparrow$0.96)} &12.89 {\color{myGreen}($\uparrow$1.46)} &9.21 {\color{myGreen}($\uparrow$0.80)} &2.23 {\color{myGreen}($\uparrow$1.21)} &0.52 {\color{myGreen}($\uparrow$0.35)} &0.21 {\color{myGreen}($\uparrow$0.19)} &63.14 {\color{myGreen}($\uparrow$1.16)} \\

    & \cellcolor{blue!20} \textbf{T5$_{\text{\model}}$} & \cellcolor{blue!20} 13.61 {\color{myGreen}($\uparrow$1.74)} & \cellcolor{blue!20} 2.03 {\color{myGreen}($\uparrow$1.02)} & \cellcolor{blue!20} 13.92 {\color{myGreen}($\uparrow$2.49)} & \cellcolor{blue!20} 9.78 {\color{myGreen}($\uparrow$1.37)} & \cellcolor{blue!20} 2.62 {\color{myGreen}($\uparrow$1.60)} & \cellcolor{blue!20} 0.51 {\color{myGreen}($\uparrow$0.34)} & \cellcolor{blue!20} 0.26 {\color{myGreen}($\uparrow$0.24)} & \cellcolor{blue!20} 63.87 {\color{myGreen}($\uparrow$1.89)} \\ \cmidrule{2-10} 

    &\textbf{BART$_{\text{SC}}$} &13.05 {\color{myGreen}($\uparrow$0.11)} &1.89 {\color{myGreen}($\uparrow$0.23)} &12.64 {\color{myGreen}($\uparrow$0.30)} &9.84 {\color{myGreen}($\uparrow$0.18)} &1.82 {\color{myGreen}($\uparrow$0.18)} &0.52 {\color{myGreen}($\uparrow$0.09)} &0.12 {\color{myGreen}($\uparrow$0.05)} &63.48 {\color{myGreen}($\uparrow$0.36)} \\
    
    &\textbf{BART$_{\text{DPA}}$} &13.12 {\color{myGreen}($\uparrow$0.18)} &1.98 {\color{myGreen}($\uparrow$0.32)} &12.81 {\color{myGreen}($\uparrow$0.47)} &9.96 {\color{myGreen}($\uparrow$0.30)} &1.94 {\color{myGreen}($\uparrow$0.30)} &0.54 {\color{myGreen}($\uparrow$0.11)} &0.15 {\color{myGreen}($\uparrow$0.08)} &63.82 {\color{myGreen}($\uparrow$0.70)}\\
    
    &\textbf{BART$_{\text{\model}-\text{Axial}}$} &13.97 {\color{myGreen}($\uparrow$1.03)} &2.21 {\color{myGreen}($\uparrow$0.55)} &13.05 {\color{myGreen}($\uparrow$0.71)} &10.16 {\color{myGreen}($\uparrow$0.50)} &2.07 {\color{myGreen}($\uparrow$0.43)} &0.61 {\color{myGreen}($\uparrow$0.18)} &0.18 {\color{myGreen}($\uparrow$0.11)} &64.34 {\color{myGreen}($\uparrow$1.22)} \\
    
    &\textbf{BART$_{\text{OT}}$} &14.29 {\color{myGreen}($\uparrow$1.35)} &2.54 {\color{myGreen}($\uparrow$0.88)} &13.72 {\color{myGreen}($\uparrow$1.38)} &10.59 {\color{myGreen}($\uparrow$0.93)} &2.16 {\color{myGreen}($\uparrow$0.52)} &0.73 {\color{myGreen}($\uparrow$0.30)} &0.22 {\color{myGreen}($\uparrow$0.15)} &65.05 {\color{myGreen}($\uparrow$1.93)} \\
    
    & \cellcolor{blue!20} \textbf{BART$_{\text{\model}}$} & \cellcolor{blue!20} \textbf{14.67} {\color{myGreen}($\uparrow$1.73)} & \cellcolor{blue!20} \textbf{2.77} {\color{myGreen}($\uparrow$1.11)} & \cellcolor{blue!20} \textbf{14.11} {\color{myGreen}($\uparrow$1.77)} & \cellcolor{blue!20} \textbf{10.92} {\color{myGreen}($\uparrow$1.26)} & \cellcolor{blue!20} 2.51 {\color{myGreen}($\uparrow$0.87)} & \cellcolor{blue!20} \textbf{0.73} {\color{myGreen}($\uparrow$0.30)} & \cellcolor{blue!20} \textbf{0.27} {\color{myGreen}($\uparrow$0.20)} & \cellcolor{blue!20} \textbf{65.93} {\color{myGreen}($\uparrow$2.81)} \\
    \bottomrule
    \end{tabular}
}
\caption{Experimental and ablation results for the response generation task with and without fusing personalities. Refer to Appendix \ref{sec:results_vis} for visualisation (Abbr: R1/2/L: ROUGE-1/2/L, B1/2/3/4: BLEU-1/2/3/4, BS: BERTScore, SC: Simple Concat, DPA: Dot Product Attention, OT: Only Traits, PA3: Personality-Aware Axial Attention).}
\label{tab:results}
\vspace{-3mm}
\end{table*}

\subsection{Response Generation}
Here, we discuss the effect of adding personality information to the dialogue context quantitatively.

\subsubsection{Comparative Systems}
To attain the most promising textual representations for discourse, we employ a range of well-established encoder-decoder-based sequence-to-sequence (seq2seq) models. (i) \textbf{RNN}: We leverage the RNN seq2seq architecture, implemented through openNMT4\footnote{\url{https://github.com/OpenNMT/OpenNMT-py}}. (ii) \textbf{Pointer Generator Network (PGN)} \citep{see2017get}: In this seq2seq architecture, a fusion of generative and copy mechanisms is harnessed, offering a versatile approach to content generation. (iii) \textbf{Transformer} \citep{vaswani2017attention}: Responses are generated using the conventional Transformer encoder-decoder model. (iv) \textbf{T5} \cite{10.5555/3455716.3455856}: We deploy the base version of the text-to-text-transfer-transformer (T5), which excels in framing multiple NLP tasks as text-to-text challenges, facilitating a unified and efficient approach to tasks such as translation, summarization, and question answering. (v) \textbf{BART} \citep{lewis2019bart}: We utilize the basic denoising autoencoder model with a bidirectional encoder and a left-to-right auto-regressive decoder.
(vi) \textbf{mBART} \citep{liu2020multilingual}: mBART\footnote{\url{https://huggingface.co/facebook/mbart-large-50-many-to-many-mmt}}, trained on multiple extensive monolingual datasets, shares the same objective and architectural structure as BART.
\subsubsection{Quantitative Results}
Table \ref{tab:results} presents the scores achieved across the evaluation metrics for the MaSaC dataset. Apparently, the inclusion of personality information elevates the performance of our comparative systems across all metrics. Notably, BART outperforms the competition, whether with or without personality information, across majority of the metrics. We observe increased ROUGE-1 scores for all models, typically ranging from $+13\%$ to $+21\%$. BLEU-1 also increases simultaneously from $+12\%$ to $+38\%$. The consistent improvement in BERTScore ($+3\%$ to $+5\%$) also underscores that the fusion of personality information into the dialogue context results in responses marked by enhanced coherence. 

\begin{figure}
    \centering
    \includegraphics[width=\columnwidth]{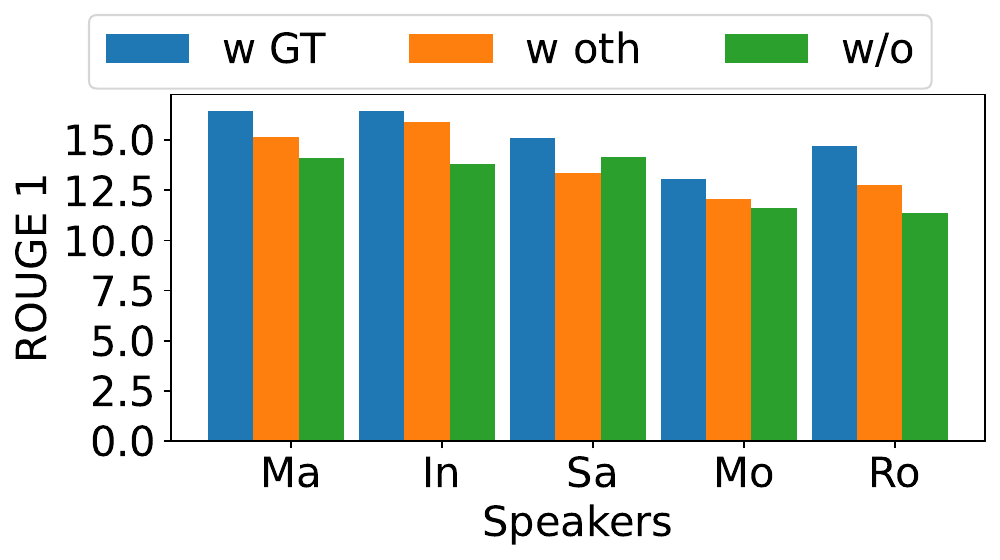}
    \caption{ROUGE-1 scores for the responses generated by the most frequent five speakers in the dataset when the GT personality, other personalities sans GT, and no personalities are used for response generation.}
    \label{fig:personaR1}
    \vspace{-4mm}
\end{figure}

\begin{table*}[ht]\centering
\resizebox{\textwidth}{!}{
    \begin{tabular}{p{8em}|p{40em}r}\toprule
    \textbf{Dialogue Context} &\textbf{Monisha:} Mummy ji, use apni pehli anniversary yaad hai, ye sunkar aap chaunki kyun? {\color{blue}\textit{(Mother-in-law, why did you seem surprised when he remembered his first anniversary?)}}\\
    
    & \textbf{Maya:} Kyunki mardon ko, cricket match kab hai, WWF kab dikhane wale hain, inn do baaton ko chhod bahut jyada yaad ni rehta. And that's fine. {\color{blue}\textit{(Because men tend to remember when the next cricket match or WWE show is rather than things like anniversaries. And that's fine.)}}\\
    
    & \textbf{Monisha:} Kya fine? Mujhe to har important din yaad rehta hai. {\color{blue}\textit{(What do you mean, fine? I remember every important day.)}}\\
    
    & \textbf{Maya:} Haan, kyunki mard aur aurat alag-alag hain. Unke dimag alag tarah se kaam krte hain. {\color{blue}\textit{(Yes, because men and women are different. Their minds work in different ways.)}} \\ \midrule
    
    \textbf{GT Response} &\textbf{Monisha:} Mai nhi manti. pichhle saal raat ko barah baje daddy ji ne aapko anniversary wish kiya tha na? Yaad krk? {\color{blue}\textit{(I don't believe that. Didn't Daddy wish you a happy anniversary last year at midnight? Remember?)}} \\ \midrule
    
    \textbf{BART} &\textbf{Monisha:} daddy ji ka dimag alag tarah se kaam krte hain. {\color{blue}\textit{(Daddy's mind work in different way.)}} \\ \midrule
    
    \textbf{BART$_{\text{\model}}$ \textit{(\textsc{Opn}})} &\textbf{Monisha:} Mai nhi manti. daddy ji ko apni pehli anniversary yaad hai. {\color{blue}\textit{(I don't believe that. Daddy remembers his first anniversary.)}} \\
    \bottomrule
    \end{tabular}
}
\caption{Responses generated for a sample dialogue from the test set of MaSaC by different model architectures.}
\label{tab:quality_results}
\vspace{-3mm}
\end{table*}
\begin{table}[t]\centering
\resizebox{\columnwidth}{!}{
    \begin{tabular}{l|lllp{5em}r}\toprule
    
    \textbf{Model} &\textbf{Fluency} &\textbf{Coherence} &\textbf{Relevancy} &\textbf{Personality oriented} \\\midrule
    
    \textbf{T5} &2.13 &2.07 &1.64 &2.01 \\
    
    \textbf{BART} &2.17 &2.03 &1.79 &2.04 \\
    
    \textbf{T5$_{\text{\model}}$} &3.07 &2.84 &2.26 &3.11 \\
    
    \cellcolor{blue!20}\textbf{BART$_{\text{\model}}$} &\cellcolor{blue!20}\textbf{3.14} &\cellcolor{blue!20}\textbf{3.09} &\cellcolor{blue!20}\textbf{2.98} &\cellcolor{blue!20}\textbf{3.23} \\
    
    \bottomrule
    \end{tabular}
}
\caption{Results of human evaluation for the response generation task.}
\label{tab:human_eval}
\vspace{-5mm}
\end{table}

\subsubsection{Effect of Personality}
We monitor ROUGE scores for responses from the top five most frequent speakers, as shown in Figure \ref{fig:data_sp_dist}. Comparing the responses generated by the BART model with ground-truth (GT) personalities (as listed in Table \ref{tab:pred_personality}), we also assess results without personality fusion. The findings, presented as ROUGE-1 scores in Figure \ref{fig:personaR1}, consistently demonstrate improved performance after personality fusion. Notably, except for Sahil, every speaker exhibits enhanced performance when infused with the GT personality within the dialogue context.

\subsubsection{Ablation Study}
It is essential to recognize that integrating personality information into the dialogue context can be achieved through various techniques, each varying in complexity. In our study, we have delved into several fusion methodologies, encompassing straightforward concatenation, conventional dot-product attention, and personality-aware attention, both with and without the inclusion of Axial attention. We provide results for both BART and T5 since they exhibit comparable capabilities in Table \ref{tab:results}. Evidently, the fusion of personality information contributes to better responses. Nevertheless, our findings emphasize that simple concatenation falls short in efficiency, yielding only marginal performance gains. In contrast, introducing attention mechanisms elevates performance, with our proposed approach of personality-aware fusion, coupled with Axial attention, being the most effective strategy. Additionally, we investigate the potential impact of fusing solely the identified personality trait without the intermediary step of mapping it into a trait definition. Our observations underscore the advantages of incorporating the complete trait definition rather than merely the isolated trait string within the response generation pipeline.

\subsubsection{Qualitative Analysis}
We select a sample dialogue from the test set and present the predicted responses generated by the conventional BART model alongside those generated after the integration of personality factors using \model. These responses are compared with the ground-truth responses, comprehensively detailed in Table \ref{tab:quality_results}. We observe that utilising personality information (\textsc{Opn} for the speaker in this case) aligns the response closer to the ground truth when compared with the standard BART model.

\subsubsection{Human Evaluation}
For generative tasks such as response generation, simple reliance on quantitative results proves insufficient, primarily due to the tendency of such metrics, like ROUGE and BLEU scores, to prioritize syntactic similarity over semantic equivalence. Therefore, we perform human evaluation. We conduct a comparative analysis of predictions derived from BART and T5 with and without the incorporation of personality information using \model. We engage $25$ human evaluators\footnote{The evaluators are linguists fluent in English and Hindi with a good grasp of personalized dialogues, aged between 25-30.} who are tasked with assessing a randomly selected set of $50$ responses generated by these methods. They assign each response a rating within the range of $1$ to $5$, considering common human evaluation metrics, including fluency, relevance, coherence, and personality orientation. Detailed definitions for each of these attributes can be found in Appendix \ref{sec:human_eval_appendix}.

To monitor the validity of the human evaluations, we calculate Cohen's Kappa \cite{mchugh2012interrater} to quantify the inter-annotator agreement between the annotators. The average Kappa score for fluency, coherence, relevancy and personality oriented came out to be $0.83$, $0.79$, $0.68$, and $0.71$, respectively. The consolidated results of our human evaluation, shown in Table \ref{tab:human_eval}, reflect the averaged ratings across all obtained responses. Evidently, BART, when equipped with personality information using \model, emerges as the top performer across all metrics.

\section{Conclusion}
We explored the task of utilising speaker personalities to aid response generation in the domain of code-mixed dialogues. Speaker personalities, from the big five personality traits, are learned in an unsupervised manner an incorporated with dialogue context using a novel fusion mechanism. We leverage a two-level attention mechanism employing context aware and Axial attention approaches to efficiently fuse the personality information with dialogue context. Our experiments demonstrated a notable improvement in response quality and coherence when personality information is fused into the systems. Furthermore, we provided insights into the inferred personality traits and their qualitative connection to response generation.

\newpage
\section{Limitations}
The study does encounter certain limitations that warrant consideration. First, the scarcity of datasets containing multiple dialogues with similar speakers in the code-mixed community limited the study to using a single dataset. While the results show promising outcomes, an investigation with multiple code-mixed datasets can also be beneficial to the community. Additionally, the dataset's source, being from a TV series lacks a real-life-like character development, introducing the possibility of inherent bias. These potential limitation highlights the need for diverse and well-rounded datasets that encompass a variety of conversational scenarios and speaker profiles to ensure the model's applicability across a broader spectrum of code-mixing instances.

\section{Ethical Considerations}
The study's ethical considerations are well-addressed in several aspects. First, the dataset used in the study is open-sourced and ethically sourced, ensuring that the data collection process adheres to ethical guidelines and data protection regulations. Second, all human annotators and evaluators involved in the research were fairly compensated for their efforts, which is a crucial ethical practice in research involving human participants. Lastly, the study poses no potential concerns related to privacy and consent, as it does not involve the collection or utilization of personal information without explicit permission. These ethical practices help maintain the integrity of the research and ensure that it aligns with ethical standards and principles.

\bibliography{custom}

\begin{thebibliography}{61}
\expandafter\ifx\csname natexlab\endcsname\relax\def\natexlab#1{#1}\fi

\bibitem[{Agarwal et~al.(2021)Agarwal, Rao, and Jayagopi}]{agarwal-etal-2021-towards}
Vibhav Agarwal, Pooja Rao, and Dinesh~Babu Jayagopi. 2021.
\newblock \href {https://doi.org/10.18653/v1/2021.nlp4convai-1.26} {Towards code-mixed {H}inglish dialogue generation}.
\newblock In \emph{Proceedings of the 3rd Workshop on Natural Language Processing for Conversational AI}, pages 271--280, Online. Association for Computational Linguistics.

\bibitem[{Ahn et~al.(2020)Ahn, Jimenez, Tsvetkov, and Black}]{ahn2020code}
Emily Ahn, Cecilia Jimenez, Yulia Tsvetkov, and Alan~W Black. 2020.
\newblock What code-switching strategies are effective in dialog systems?
\newblock In \emph{Proceedings of the Society for Computation in Linguistics 2020}, pages 254--264.

\bibitem[{Alam and Riccardi(2014)}]{alam2014fusion}
Firoj Alam and Giuseppe Riccardi. 2014.
\newblock Fusion of acoustic, linguistic and psycholinguistic features for speaker personality traits recognition.
\newblock In \emph{2014 IEEE international conference on acoustics, speech and signal processing (ICASSP)}, pages 955--959. IEEE.

\bibitem[{Ameer et~al.(2022)Ameer, Sidorov, Gomez-Adorno, and Nawab}]{ameer2022multi}
Iqra Ameer, Grigori Sidorov, Helena Gomez-Adorno, and Rao Muhammad~Adeel Nawab. 2022.
\newblock Multi-label emotion classification on code-mixed text: Data and methods.
\newblock \emph{IEEE Access}, 10:8779--8789.

\bibitem[{Banerjee et~al.(2018)Banerjee, Moghe, Arora, and Khapra}]{banerjee-etal-2018-dataset}
Suman Banerjee, Nikita Moghe, Siddhartha Arora, and Mitesh~M. Khapra. 2018.
\newblock \href {https://aclanthology.org/C18-1319} {A dataset for building code-mixed goal oriented conversation systems}.
\newblock In \emph{Proceedings of the 27th International Conference on Computational Linguistics}, pages 3766--3780, Santa Fe, New Mexico, USA. Association for Computational Linguistics.

\bibitem[{Bawa et~al.(2020)Bawa, Khadpe, Joshi, Bali, and Choudhury}]{10.1145/3392846}
Anshul Bawa, Pranav Khadpe, Pratik Joshi, Kalika Bali, and Monojit Choudhury. 2020.
\newblock \href {https://doi.org/10.1145/3392846} {Do multilingual users prefer chat-bots that code-mix? let's nudge and find out!}
\newblock \emph{Proc. ACM Hum.-Comput. Interact.}, 4(CSCW1).

\bibitem[{Bedi et~al.(2021)Bedi, Kumar, Akhtar, and Chakraborty}]{bedi2021multi}
Manjot Bedi, Shivani Kumar, Md~Shad Akhtar, and Tanmoy Chakraborty. 2021.
\newblock Multi-modal sarcasm detection and humor classification in code-mixed conversations.
\newblock \emph{IEEE Transactions on Affective Computing}.

\bibitem[{Benjamin~Jr(2020)}]{inbook}
Arlin Benjamin~Jr. 2020.
\newblock \href {https://doi.org/10.1002/9781118970843.ch328} {\emph{Type A/B Personalities}}.

\bibitem[{Briggs and Myers(1995)}]{myers1995}
Myers~Isabel Briggs and Peter~B. Myers. 1995.
\newblock \emph{Gifts Differing : Understanding Personality Type}.
\newblock Davies-Black Publishing.

\bibitem[{Bukhari et~al.(2023)Bukhari, Zubair, and Arshad}]{bukhari2023humor}
Syed Husnain~Haider Bukhari, Anusha Zubair, and Muhammad~Umair Arshad. 2023.
\newblock Humor detection in english-urdu code-mixed language.
\newblock In \emph{2023 3rd International Conference on Artificial Intelligence (ICAI)}, pages 26--31. IEEE.

\bibitem[{Butcher and Williams(2009)}]{https://doi.org/10.1111/j.1758-0854.2008.01007.x}
James~N. Butcher and Carolyn~L. Williams. 2009.
\newblock \href {https://doi.org/https://doi.org/10.1111/j.1758-0854.2008.01007.x} {Personality assessment with the mmpi-2: Historical roots, international adaptations, and current challenges}.
\newblock \emph{Applied Psychology: Health and Well-Being}, 1(1):105--135.

\bibitem[{Celli and Lepri(2018)}]{celli2018big}
Fabio Celli and Bruno Lepri. 2018.
\newblock Is big five better than mbti? a personality computing challenge using twitter data.
\newblock In \emph{CLiC-it}.

\bibitem[{Chen et~al.(2020{\natexlab{a}})Chen, Zheng, and Du}]{chen-etal-2020-listeners}
Guanyi Chen, Yinhe Zheng, and Yupei Du. 2020{\natexlab{a}}.
\newblock \href {https://aclanthology.org/2020.inlg-1.26} {Listener{'}s social identity matters in personalised response generation}.
\newblock In \emph{Proceedings of the 13th International Conference on Natural Language Generation}, pages 205--215, Dublin, Ireland. Association for Computational Linguistics.

\bibitem[{Chen et~al.(2020{\natexlab{b}})Chen, Zheng, and Du}]{chen2020listener}
Guanyi Chen, Yinhe Zheng, and Yupei Du. 2020{\natexlab{b}}.
\newblock Listener's social identity matters in personalised response generation.
\newblock \emph{arXiv preprint arXiv:2010.14342}.

\bibitem[{Chen et~al.(2017)Chen, Liu, Yin, and Tang}]{10.1145/3166054.3166058}
Hongshen Chen, Xiaorui Liu, Dawei Yin, and Jiliang Tang. 2017.
\newblock \href {https://doi.org/10.1145/3166054.3166058} {A survey on dialogue systems: Recent advances and new frontiers}.
\newblock \emph{SIGKDD Explor. Newsl.}, 19(2):25–35.

\bibitem[{Costa and McCrae(1992)}]{costa1992normal}
Paul~T Costa and Robert~R McCrae. 1992.
\newblock Normal personality assessment in clinical practice: The neo personality inventory.
\newblock \emph{Psychological assessment}, 4(1):5.

\bibitem[{Costa~Jr and McCrae(2008)}]{costa2008revised}
Paul~T Costa~Jr and Robert~R McCrae. 2008.
\newblock \emph{The Revised Neo Personality Inventory (neo-pi-r).}
\newblock Sage Publications, Inc.

\bibitem[{Digman(1990)}]{doi:10.1146/annurev.ps.41.020190.002221}
J~M Digman. 1990.
\newblock \href {https://doi.org/10.1146/annurev.ps.41.020190.002221} {Personality structure: Emergence of the five-factor model}.
\newblock \emph{Annual Review of Psychology}, 41(1):417--440.

\bibitem[{Dinan et~al.(2018)Dinan, Roller, Shuster, Fan, Auli, and Weston}]{dinan2018wizard}
Emily Dinan, Stephen Roller, Kurt Shuster, Angela Fan, Michael Auli, and Jason Weston. 2018.
\newblock Wizard of wikipedia: Knowledge-powered conversational agents.
\newblock \emph{arXiv preprint arXiv:1811.01241}.

\bibitem[{Dong et~al.(2022)Dong, Li, Gong, Chen, Li, Shen, and Yang}]{10.1145/3554727}
Chenhe Dong, Yinghui Li, Haifan Gong, Miaoxin Chen, Junxin Li, Ying Shen, and Min Yang. 2022.
\newblock \href {https://doi.org/10.1145/3554727} {A survey of natural language generation}.
\newblock \emph{ACM Comput. Surv.}, 55(8).

\bibitem[{Dowlagar and Mamidi(2023)}]{DOWLAGAR2023101449}
Suman Dowlagar and Radhika Mamidi. 2023.
\newblock \href {https://doi.org/https://doi.org/10.1016/j.csl.2022.101449} {A code-mixed task-oriented dialog dataset for medical domain}.
\newblock \emph{Computer Speech and Language}, 78:101449.

\bibitem[{Fan et~al.(2020)Fan, Luo, and Lin}]{fan2020survey}
Yifan Fan, Xudong Luo, and Pingping Lin. 2020.
\newblock A survey of response generation of dialogue systems.
\newblock \emph{International Journal of Computer and Information Engineering}, 14(12):461--472.

\bibitem[{Firdaus et~al.(2023)Firdaus, Ekbal, and Cambria}]{FIRDAUS2023299}
Mauajama Firdaus, Asif Ekbal, and Erik Cambria. 2023.
\newblock \href {https://doi.org/https://doi.org/10.1016/j.inffus.2022.09.029} {Multitask learning for multilingual intent detection and slot filling in dialogue systems}.
\newblock \emph{Information Fusion}, 91:299--315.

\bibitem[{Golbeck et~al.(2011)Golbeck, Robles, and Turner}]{golbeck2011predicting}
Jennifer Golbeck, Cristina Robles, and Karen Turner. 2011.
\newblock Predicting personality with social media.
\newblock In \emph{CHI'11 extended abstracts on human factors in computing systems}, pages 253--262.

\bibitem[{Gottardi et~al.(2022)Gottardi, Ipek, Castellucci, Hu, Vaz, Lu, Khatri, Chadha, Zhang, Sahai, Dwivedi, Shi, Hu, Huang, Dai, Yang, Somani, Rajan, Rezac, and Maarek}]{alexabot}
Anna Gottardi, Osman Ipek, Giuseppe Castellucci, Shui Hu, Lavina Vaz, Yao Lu, Anju Khatri, Anjali Chadha, Desheng Zhang, Sattvik Sahai, Prerna Dwivedi, Hangjie Shi, Lucy Hu, Andy Huang, Luke Dai, Bofei Yang, Varun Somani, Pankaj Rajan, Ron Rezac, and Yoelle Maarek. 2022.
\newblock \href {https://doi.org/10.48550/arXiv.2209.06321} {Alexa, let's work together: Introducing the first alexa prize taskbot challenge on conversational task assistance}.

\bibitem[{Ho et~al.(2020)Ho, Kalchbrenner, Weissenborn, and Salimans}]{ho2020axial}
Jonathan Ho, Nal Kalchbrenner, Dirk Weissenborn, and Tim Salimans. 2020.
\newblock \href {https://openreview.net/forum?id=H1e5GJBtDr} {Axial attention in multidimensional transformers}.

\bibitem[{Joshi et~al.(2017)Joshi, Mi, and Faltings}]{joshi2017personalization}
Chaitanya~K Joshi, Fei Mi, and Boi Faltings. 2017.
\newblock Personalization in goal-oriented dialog.
\newblock \emph{arXiv preprint arXiv:1706.07503}.

\bibitem[{Kasper and Wagner(2014)}]{kasper_wagner_2014}
Gabriele Kasper and Johannes Wagner. 2014.
\newblock \href {https://doi.org/10.1017/S0267190514000014} {Conversation analysis in applied linguistics}.
\newblock \emph{Annual Review of Applied Linguistics}, 34:171–212.

\bibitem[{Khandelwal et~al.(2018)Khandelwal, Swami, Akhtar, and Shrivastava}]{khandelwal-etal-2018-humor}
Ankush Khandelwal, Sahil Swami, Syed~S. Akhtar, and Manish Shrivastava. 2018.
\newblock \href {https://aclanthology.org/L18-1193} {Humor detection in {E}nglish-{H}indi code-mixed social media content : Corpus and baseline system}.
\newblock In \emph{Proceedings of the Eleventh International Conference on Language Resources and Evaluation ({LREC} 2018)}, Miyazaki, Japan. European Language Resources Association (ELRA).

\bibitem[{Kosinski et~al.(2013)Kosinski, Stillwell, and Graepel}]{kosinski2013private}
Michal Kosinski, David Stillwell, and Thore Graepel. 2013.
\newblock Private traits and attributes are predictable from digital records of human behavior.
\newblock \emph{Proceedings of the national academy of sciences}, 110(15):5802--5805.

\bibitem[{Kumar et~al.(2023{\natexlab{a}})Kumar, Bhatia, Aggarwal, and Chakraborty}]{kumar2023dialogue}
Shivani Kumar, Sumit Bhatia, Milan Aggarwal, and Tanmoy Chakraborty. 2023{\natexlab{a}}.
\newblock \href {http://arxiv.org/abs/2307.07255} {Dialogue agents 101: A beginner's guide to critical ingredients for designing effective conversational systems}.

\bibitem[{Kumar et~al.(2022)Kumar, Kulkarni, Akhtar, and Chakraborty}]{kumar-etal-2022-become}
Shivani Kumar, Atharva Kulkarni, Md~Shad Akhtar, and Tanmoy Chakraborty. 2022.
\newblock \href {https://doi.org/10.18653/v1/2022.acl-long.411} {When did you become so smart, oh wise one?! sarcasm explanation in multi-modal multi-party dialogues}.
\newblock In \emph{Proceedings of the 60th Annual Meeting of the Association for Computational Linguistics (Volume 1: Long Papers)}, pages 5956--5968, Dublin, Ireland. Association for Computational Linguistics.

\bibitem[{Kumar et~al.(2023{\natexlab{b}})Kumar, Mondal, Akhtar, and Chakraborty}]{kumar2023explaining}
Shivani Kumar, Ishani Mondal, Md~Shad Akhtar, and Tanmoy Chakraborty. 2023{\natexlab{b}}.
\newblock Explaining (sarcastic) utterances to enhance affect understanding in multimodal dialogues.
\newblock In \emph{Proceedings of the AAAI Conference on Artificial Intelligence}, volume~37, pages 12986--12994.

\bibitem[{Lewis et~al.(2020)Lewis, Liu, Goyal, Ghazvininejad, Mohamed, Levy, Stoyanov, and Zettlemoyer}]{lewis2019bart}
Mike Lewis, Yinhan Liu, Naman Goyal, Marjan Ghazvininejad, Abdelrahman Mohamed, Omer Levy, Veselin Stoyanov, and Luke Zettlemoyer. 2020.
\newblock \href {https://doi.org/10.18653/v1/2020.acl-main.703} {{BART}: Denoising sequence-to-sequence pre-training for natural language generation, translation, and comprehension}.
\newblock In \emph{Proceedings of the 58th Annual Meeting of the Association for Computational Linguistics}, pages 7871--7880, Online. Association for Computational Linguistics.

\bibitem[{Li et~al.(2016)Li, Galley, Brockett, Spithourakis, Gao, and Dolan}]{li2016persona}
Jiwei Li, Michel Galley, Chris Brockett, Georgios~P Spithourakis, Jianfeng Gao, and Bill Dolan. 2016.
\newblock A persona-based neural conversation model.
\newblock \emph{arXiv preprint arXiv:1603.06155}.

\bibitem[{Lin(2004)}]{lin-2004-rouge}
Chin-Yew Lin. 2004.
\newblock \href {https://aclanthology.org/W04-1013} {{ROUGE}: A package for automatic evaluation of summaries}.
\newblock In \emph{Text Summarization Branches Out}, pages 74--81, Barcelona, Spain. Association for Computational Linguistics.

\bibitem[{Liu et~al.(2020{\natexlab{a}})Liu, Gu, Goyal, Li, Edunov, Ghazvininejad, Lewis, and Zettlemoyer}]{liu2020multilingual}
Yinhan Liu, Jiatao Gu, Naman Goyal, Xian Li, Sergey Edunov, Marjan Ghazvininejad, Mike Lewis, and Luke Zettlemoyer. 2020{\natexlab{a}}.
\newblock \href {https://doi.org/10.1162/tacl_a_00343} {Multilingual denoising pre-training for neural machine translation}.
\newblock \emph{Transactions of the Association for Computational Linguistics}, 8:726--742.

\bibitem[{Liu et~al.(2020{\natexlab{b}})Liu, Ott, Goyal, Du, Joshi, Chen, Levy, Lewis, Zettlemoyer, and Stoyanov}]{liu2020roberta}
Yinhan Liu, Myle Ott, Naman Goyal, Jingfei Du, Mandar Joshi, Danqi Chen, Omer Levy, Mike Lewis, Luke Zettlemoyer, and Veselin Stoyanov. 2020{\natexlab{b}}.
\newblock \href {https://openreview.net/forum?id=SyxS0T4tvS} {Ro{\{}bert{\}}a: A robustly optimized {\{}bert{\}} pretraining approach}.

\bibitem[{Liu et~al.(2020{\natexlab{c}})Liu, Winata, Lin, Xu, and Fung}]{Liu_Winata_Lin_Xu_Fung_2020}
Zihan Liu, Genta~Indra Winata, Zhaojiang Lin, Peng Xu, and Pascale Fung. 2020{\natexlab{c}}.
\newblock \href {https://doi.org/10.1609/aaai.v34i05.6362} {Attention-informed mixed-language training for zero-shot cross-lingual task-oriented dialogue systems}.
\newblock \emph{Proceedings of the AAAI Conference on Artificial Intelligence}, 34(05):8433--8440.

\bibitem[{Lucas et~al.(2009)Lucas, Fern{\'a}ndez, Salazar, Ferreiros, and San~Segundo}]{lucas2009managing}
JM~Lucas, F~Fern{\'a}ndez, J~Salazar, J~Ferreiros, and R~San~Segundo. 2009.
\newblock Managing speaker identity and user profiles in a spoken dialogue system.
\newblock \emph{Procesamiento del lenguaje natural}, (43):77--84.

\bibitem[{Madhu et~al.(2023)Madhu, Satapara, Modha, Mandl, and Majumder}]{madhu2023detecting}
Hiren Madhu, Shrey Satapara, Sandip Modha, Thomas Mandl, and Prasenjit Majumder. 2023.
\newblock Detecting offensive speech in conversational code-mixed dialogue on social media: A contextual dataset and benchmark experiments.
\newblock \emph{Expert Systems with Applications}, 215:119342.

\bibitem[{Mairesse et~al.(2007)Mairesse, Walker, Mehl, and Moore}]{mairesse2007using}
Fran{\c{c}}ois Mairesse, Marilyn~A Walker, Matthias~R Mehl, and Roger~K Moore. 2007.
\newblock Using linguistic cues for the automatic recognition of personality in conversation and text.
\newblock \emph{Journal of artificial intelligence research}, 30:457--500.

\bibitem[{McHugh(2012)}]{mchugh2012interrater}
Mary~L McHugh. 2012.
\newblock Interrater reliability: the kappa statistic.
\newblock \emph{Biochemia medica}, 22(3):276--282.

\bibitem[{Mikolov et~al.(2013)Mikolov, Chen, Corrado, and Dean}]{mikolov2013efficient}
Tomas Mikolov, Kai Chen, Greg Corrado, and Jeffrey Dean. 2013.
\newblock Efficient estimation of word representations in vector space.
\newblock \emph{arXiv preprint arXiv:1301.3781}.

\bibitem[{Modha et~al.(2021)Modha, Mandl, Shahi, Madhu, Satapara, Ranasinghe, and Zampieri}]{modha2021overview}
Sandip Modha, Thomas Mandl, Gautam~Kishore Shahi, Hiren Madhu, Shrey Satapara, Tharindu Ranasinghe, and Marcos Zampieri. 2021.
\newblock Overview of the hasoc subtrack at fire 2021: Hate speech and offensive content identification in english and indo-aryan languages and conversational hate speech.
\newblock In \emph{Proceedings of the 13th Annual Meeting of the Forum for Information Retrieval Evaluation}, pages 1--3.

\bibitem[{Papineni et~al.(2002)Papineni, Roukos, Ward, and Zhu}]{papineni2002bleu}
Kishore Papineni, Salim Roukos, Todd Ward, and Wei-Jing Zhu. 2002.
\newblock \href {https://doi.org/10.3115/1073083.1073135} {{B}leu: a method for automatic evaluation of machine translation}.
\newblock In \emph{Proceedings of the 40th Annual Meeting of the Association for Computational Linguistics}, pages 311--318, Philadelphia, Pennsylvania, USA. Association for Computational Linguistics.

\bibitem[{Raffel et~al.(2020)Raffel, Shazeer, Roberts, Lee, Narang, Matena, Zhou, Li, and Liu}]{10.5555/3455716.3455856}
Colin Raffel, Noam Shazeer, Adam Roberts, Katherine Lee, Sharan Narang, Michael Matena, Yanqi Zhou, Wei Li, and Peter~J. Liu. 2020.
\newblock Exploring the limits of transfer learning with a unified text-to-text transformer.
\newblock \emph{J. Mach. Learn. Res.}, 21(1).

\bibitem[{Roller et~al.(2020)Roller, Dinan, Goyal, Ju, Williamson, Liu, Xu, Ott, Shuster, Smith et~al.}]{roller2020recipes}
Stephen Roller, Emily Dinan, Naman Goyal, Da~Ju, Mary Williamson, Yinhan Liu, Jing Xu, Myle Ott, Kurt Shuster, Eric~M Smith, et~al. 2020.
\newblock Recipes for building an open-domain chatbot.
\newblock \emph{arXiv preprint arXiv:2004.13637}.

\bibitem[{Schwartz et~al.(2013)Schwartz, Eichstaedt, Kern, Dziurzynski, Ramones, Agrawal, Shah, Kosinski, Stillwell, Seligman et~al.}]{schwartz2013personality}
H~Andrew Schwartz, Johannes~C Eichstaedt, Margaret~L Kern, Lukasz Dziurzynski, Stephanie~M Ramones, Megha Agrawal, Achal Shah, Michal Kosinski, David Stillwell, Martin~EP Seligman, et~al. 2013.
\newblock Personality, gender, and age in the language of social media: The open-vocabulary approach.
\newblock \emph{PloS one}, 8(9):e73791.

\bibitem[{See et~al.(2017)See, Liu, and Manning}]{see2017get}
Abigail See, Peter~J. Liu, and Christopher~D. Manning. 2017.
\newblock \href {https://doi.org/10.18653/v1/P17-1099} {Get to the point: Summarization with pointer-generator networks}.
\newblock In \emph{Proceedings of the 55th Annual Meeting of the Association for Computational Linguistics (Volume 1: Long Papers)}, pages 1073--1083, Vancouver, Canada. Association for Computational Linguistics.

\bibitem[{Singh et~al.(2022)Singh, Firdaus, Shambhavi, Mishra, and Ekbal}]{SINGH2022108900}
Gopendra~Vikram Singh, Mauajama Firdaus, Shambhavi, Shruti Mishra, and Asif Ekbal. 2022.
\newblock \href {https://doi.org/https://doi.org/10.1016/j.knosys.2022.108900} {Knowing what to say: Towards knowledge grounded code-mixed response generation for open-domain conversations}.
\newblock \emph{Knowledge-Based Systems}, 249:108900.

\bibitem[{Spring et~al.(2019)Spring, Casas, Daher, Mugellini, and Abou~Khaled}]{spring2019empathic}
Timo Spring, Jacky Casas, Karl Daher, Elena Mugellini, and Omar Abou~Khaled. 2019.
\newblock Empathic response generation in chatbots.
\newblock In \emph{Proceedings of 4th Swiss Text Analytics Conference (SwissText 2019), 18-19 June 2019, Wintherthur, Switzerland}. 18-19 June 2019.

\bibitem[{Tarihoran and Sumirat(2022)}]{tarihoran2022impact}
Naf’an Tarihoran and Iin~Ratna Sumirat. 2022.
\newblock The impact of social media on the use of code mixing by generation z.
\newblock \emph{International Journal of Interactive Mobile Technologies (iJIM)}, 16(7):54--69.

\bibitem[{Tay(1989)}]{tay1989code}
Mary~WJ Tay. 1989.
\newblock Code switching and code mixing as a communicative strategy in multilingual discourse.
\newblock \emph{World Englishes}, 8(3):407--417.

\bibitem[{Turnbull(2003)}]{turnbull2003language}
William Turnbull. 2003.
\newblock \emph{Language in action: Psychological models of conversation}.
\newblock Routledge.

\bibitem[{Vaswani et~al.(2017)Vaswani, Shazeer, Parmar, Uszkoreit, Jones, Gomez, Kaiser, and Polosukhin}]{vaswani2017attention}
Ashish Vaswani, Noam Shazeer, Niki Parmar, Jakob Uszkoreit, Llion Jones, Aidan~N Gomez, \L~ukasz Kaiser, and Illia Polosukhin. 2017.
\newblock \href {https://proceedings.neurips.cc/paper/2017/file/3f5ee243547dee91fbd053c1c4a845aa-Paper.pdf} {Attention is all you need}.
\newblock In \emph{Advances in Neural Information Processing Systems}, volume~30. Curran Associates, Inc.

\bibitem[{Weston et~al.(2018)Weston, Dinan, and Miller}]{weston2018retrieve}
Jason Weston, Emily Dinan, and Alexander~H Miller. 2018.
\newblock Retrieve and refine: Improved sequence generation models for dialogue.
\newblock \emph{arXiv preprint arXiv:1808.04776}.

\bibitem[{Yang et~al.(2019)Yang, Li, Wong, Chao, Wang, and Tu}]{Yang_Li_Wong_Chao_Wang_Tu_2019}
Baosong Yang, Jian Li, Derek~F. Wong, Lidia~S. Chao, Xing Wang, and Zhaopeng Tu. 2019.
\newblock \href {https://doi.org/10.1609/aaai.v33i01.3301387} {Context-aware self-attention networks}.
\newblock \emph{Proceedings of the AAAI Conference on Artificial Intelligence}, 33(01):387--394.

\bibitem[{Zhang et~al.(2018{\natexlab{a}})Zhang, Dinan, Urbanek, Szlam, Kiela, and Weston}]{zhang-etal-2018-personalizing}
Saizheng Zhang, Emily Dinan, Jack Urbanek, Arthur Szlam, Douwe Kiela, and Jason Weston. 2018{\natexlab{a}}.
\newblock \href {https://doi.org/10.18653/v1/P18-1205} {Personalizing dialogue agents: {I} have a dog, do you have pets too?}
\newblock In \emph{Proceedings of the 56th Annual Meeting of the Association for Computational Linguistics (Volume 1: Long Papers)}, pages 2204--2213, Melbourne, Australia. Association for Computational Linguistics.

\bibitem[{Zhang et~al.(2018{\natexlab{b}})Zhang, Dinan, Urbanek, Szlam, Kiela, and Weston}]{zhang2018personalizing}
Saizheng Zhang, Emily Dinan, Jack Urbanek, Arthur Szlam, Douwe Kiela, and Jason Weston. 2018{\natexlab{b}}.
\newblock Personalizing dialogue agents: I have a dog, do you have pets too?
\newblock \emph{arXiv preprint arXiv:1801.07243}.

\bibitem[{Zhang et~al.(2019)Zhang, Kishore, Wu, Weinberger, and Artzi}]{zhang2019bertscore}
Tianyi Zhang, Varsha Kishore, Felix Wu, Kilian~Q Weinberger, and Yoav Artzi. 2019.
\newblock \href {https://openreview.net/forum?id=SkeHuCVFDr} {Bertscore: Evaluating text generation with bert}.
\newblock In \emph{International Conference on Learning Representations}.

\end{thebibliography}

\appendix

\newpage
\section{Appendix}
\label{sec:appendix}
\subsection{Big Five Personality Traits}
\label{sec:big-5}

The widely accepted Big Five Personality Trait Model \cite{doi:10.1146/annurev.ps.41.020190.002221} is a valuable framework for understanding human personality. It consists of five core traits, abbreviated as \textsc{OCEAN} - openness, conscientiousness, extraversion, agreeableness, and neuroticism. These traits provide unique perspectives for character assessment, forming a comprehensive quantitative framework. For detailed definitions and examples of each trait, refer to Table \ref{tab:bigfivedef}.

\begin{table*}[t]
    \centering
    \resizebox{\textwidth}{!}{
    \begin{tabular}{l|p{17em}|p{17em}}
        \toprule
        \textbf{Trait} & \textbf{Definition} & \textbf{Example}\\
        \midrule
        Openness & This trait reflects a person's willingness to explore new ideas, engage in creative activities, and embrace novel experiences.
        & Someone high in openness might enjoy trying exotic cuisines, artistic endeavors, and philosophical discussions. \\ \midrule
        Conscientiousness & Conscientious individuals are organized, goal-oriented, and reliable. They tend to plan ahead and complete tasks with precision. & A conscientious person may meticulously prepare a project schedule and consistently meet deadlines. \\ \midrule
        Extraversion & Extraversion refers to the degree of sociability, assertiveness, and enthusiasm in an individual. & An extrovert is more likely to enjoy social gatherings, initiate conversations, and thrive in group settings. \\ \midrule
        Agreeableness & Agreeable individuals are characterized by their empathy, cooperativeness, and willingness to accommodate others. & An agreeable person is more likely to compromise during conflicts and be a supportive friend. \\ \midrule
        Neuroticism & Neuroticism reflects emotional stability and the tendency to experience negative emotions like anxiety and insecurity. & A highly neurotic person might often worry about various aspects of their life and react strongly to stressors. \\
        \bottomrule
    \end{tabular}}
    \caption{Definitions and Examples of the Big Five Personality Traits}
    \label{tab:bigfivedef}
\end{table*}

\subsection{Characteristics Descriptions for Dataset Speakers}
\label{sec:speaker_char}
Drawing insights from Figure \ref{fig:data_sp_dist}, we select the top five frequent speakers, namely Maya, Indravardhan, Sahil, Monisha, and Rosesh, from the extensive MaSaC dataset. These individuals are pivotal for our in-depth analysis. To validate our predicted personalities, human annotators with expertise assess the actual personalities of these speakers. To aid this evaluation, we utilize detailed character descriptions from the Wikipedia page\footnote{\url{https://en.wikipedia.org/wiki/Sarabhai_vs_Sarabhai}} of the show 'Sarabhai v/s Sarabhai'\footnote{\url{https://www.imdb.com/title/tt1518542/}}, presented in Table \ref{tab:speaker_pers}. Annotators refer to these descriptions when assigning personality traits from the big-five personality traits to each speaker.

\begin{table*}[t]
\centering
\resizebox{\textwidth}{!}{
\begin{tabular}{l|p{38em}r}\toprule
\textbf{Speaker} &\textbf{Character Description on Wikipedia} \\\midrule
Maya & Maya Sarabhai is the female head of the Sarabhai family and runs the family like a pro. Being a snooty upper-class socialite, her daughter-in-law Monisha's middle-class money-saving techniques and unkempt behavior are constant pet peeves for Maya. Her catchphrase is "It's catastrophically middle class!", and she continually uses sarcasm to taunt Monisha and make her see the error of her ways. Whenever she taunts Monisha, depending on the intensity of the taunts, one to three bullet shots are heard in the background, increasing the humor in these situations and portraying her as a verbal bullet. She is constantly after Indravadhan to fix his dietary and cleanliness habits, not much unlike Monisha, and pampers her younger son Rosesh, also making sure he doesn't take a middle-class wife like Sahil. Her son-in-law Dushyant also irritates her by dropping in every time an appliance is damaged. \\ \midrule
Indravardhan & Indravadhan Sarabhai a.k.a. Indu, is an ex-director of a multinational company. He retired early to take care of the children and help Maya work as a social worker. He is always in conflict with his youngest son, Rosesh, he also jokes with Maya, pretending to hate her but actually loving her dearly as portrayed in various episodes. He constantly picks on Maya and Rosesh, always siding with Monisha in case of a tiff between her and Maya, and constantly tries to create conflicts between them. He notoriously ignites most of the quarrels in the family and then takes the seat in the audience, enjoying himself. He is irritated by his brother-in-law Madhusudan Bhai and his "hain?", as well as Dushyant, his son-in-law. He is the jester in the family. \\ \midrule
Sahil & Sahil Sarabhai is a cosmetologist. He is the eldest child, and arguably the most normal one in his otherwise eccentric family. He is soft, calm, wise and noble, and is constantly trying to resolve conflicts in his family, between Maya and Monisha, Maya and Indravadhan and Rosesh. He often gets sandwiched between his mother and his wife and tries not to hurt anyone. He avoids conflicts but loves making fun of his younger brother Rosesh, similar to Indravadhan. \\ \midrule
Monisha & Monisha Sarabhai is a middle class, Punjabi girl from Noida and now the daughter-in-law of the Sarabhai’s. She rarely cleans the house and is always lazing around watching daily soaps on television. She develops a dramatic nature from these shows and always ends up saying threatening Sahil with leaving the house after every argument with Maya. Her passion is to save money, come what may. She is always at loggerheads with Maya for her thrifty ways. Her father-in-law always supports her, while Sahil is torn between the two. Despite being careless, Monisha is an honest, innocent, and loving woman. Manisha was named 'Monisha' by Maya as she found the name Manisha 'too middle-class'. \\ \midrule
Rosesh & Rosesh Sarabhai is the youngest child of Maya and Indravadhan. He is a theatre artist, an aspiring actor, and a so-called poet. He is Maya's favorite and she pampers him a lot. He wants to become an actor and his mother Maya supports him the most. Maya is the only member of the Sarabhai family who approves of and appreciates his absurd poetry and acting skills. He has a love-hate relationship with Indravadhan as he is always the target of his jokes and pranks. He always seconds his momma even if he doesn't feel like it. He has a peculiar and amusing voice, and his poems are always bad but funny. \\
\bottomrule
\end{tabular}}
\caption{Character definition as present on Wikipedia of the most frequent five speakers in MaSaC dataset.}
\label{tab:speaker_pers}
\end{table*}

\subsection{Human Annotations for Evaluating Personality Identification}
\label{sec:pers_id}
To validate the RoBERTa model's personality predictions for our top five speakers, we enlisted the input of five human annotators. These annotators, proficient in English and Hindi, were tasked with assigning one of the Big Five personality traits to each speaker based on character descriptions (see Table \ref{tab:speaker_pers}). Their ages range between 25-30. We assessed inter-annotator agreement using the Cohen Kappa method \cite{mchugh2012interrater}, which yielded an agreement score of 0.78, confirming the reliability of our ground truth.

\begin{figure}[t]
    \centering
    \includegraphics[width=\columnwidth]{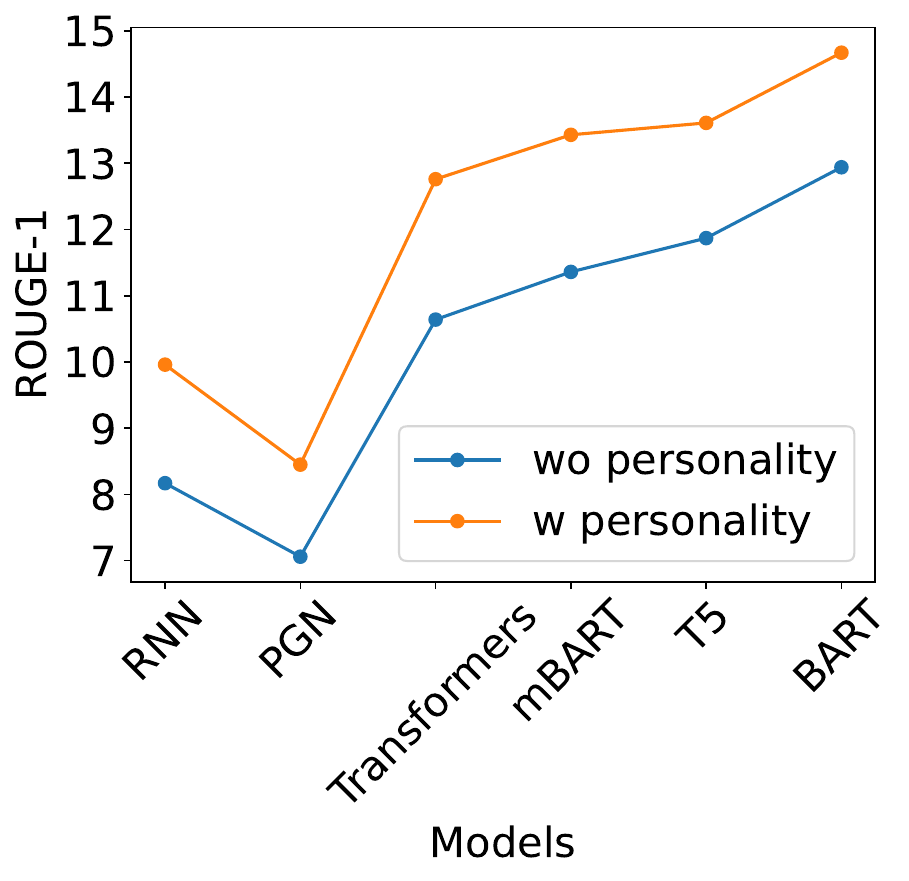}
    \caption{ROUGE-1 score visualisation shows a consistent increase in model performance when personality is infused with dialogue context.}
    \label{fig:compare}
    \vspace{-3mm}
\end{figure}

\begin{figure}[!ht]
    \centering
    \includegraphics[width=\columnwidth]{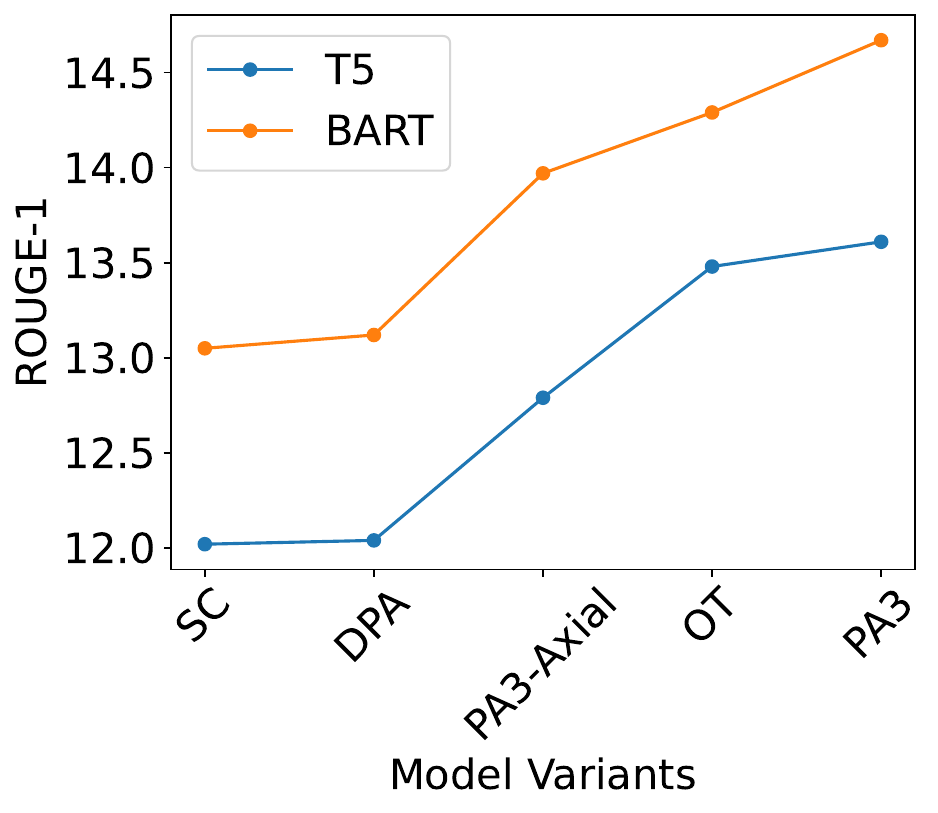}
    \caption{ROUGE-1 score visualisation shows a consistent increase in model performance when we change the fusion method. (Abbr: SC: Simple Concat, DPA: Dot Product Attention, OT: Only Traits, PA3: Personality-Aware Axial Attention).}
    \label{fig:compare2}
    \vspace{-3mm}
\end{figure}

\subsection{Visualisation of Results} \label{sec:results_vis}
In this section, we visualise the ROUGE-1 scores that we obtain for the task of response generation from the standard models without fusing personalities and after fusing personalities using \model\. Figure \ref{fig:compare} illustrates these findings. It can clearly be observed that there is a consistent increase in the response generation performance when personality is fused into the system for all models. Additionally, we also visualise the increase in performance when we increase the fusion efficiency by ranging the fusion mechanism from simple concat to the proposed \model\ in Figure \ref{fig:compare2}.

\subsection{Human Evaluation} \label{sec:human_eval_appendix}
For generative tasks like response generation, quantitative metrics alone may not offer a complete evaluation, as they tend to favor syntactic similarity over semantic equivalence. To address this, we utilize human evaluation to provide a more comprehensive assessment. Our approach considers key characteristics to gain a deeper understanding of response quality:

\begin{itemize}[leftmargin=*,noitemsep,topsep=0pt]
    \item \textbf{Fluency:} This dimension assesses the naturalness and readability of the generated text. It focuses on grammar, syntax, and language flow, with higher scores indicating smoother and more linguistically proficient text.
    \item \textbf{Relevance:} The relevance aspect measures how effectively the generated text aligns with the given context or prompt. It evaluates the appropriateness of content in relation to the context, with higher scores signifying a stronger alignment between the response and the context.
    \item \textbf{Coherence:} Coherence evaluation pertains to the logical flow and semantic connection of ideas within the generated text. It ensures that the information is well-structured, logically connected, and readily comprehensible. Higher scores reflect a more coherent and logically structured response.
    \item \textbf{Relevance to Personality:} This specific dimension evaluates whether the generated response is pertinent to the target speaker's personality. It is a crucial element in our evaluation, as it directly relates to the effectiveness of incorporating personality traits into the generated text.
\end{itemize}

This comprehensive approach offers a nuanced assessment of response generation quality, enhancing our understanding of the system's performance in language, context, and personality capture. See Table \ref{tab:human_eval} for the summarized evaluation results.

\subsection{Training System and Hyperparameter Tuning}
We mention below the computational framework we use to train our models.
\begin{itemize}[leftmargin=*,noitemsep,topsep=0pt]
    \item Description of computing infrastructure used
        \begin{itemize}
            \item Linux $64$ Bit
            \item GPU: Tesla-V100 ($32510$ MiB)
        \end{itemize}
    \item Trainable parameter: $326368976$
    \item Average runtime: 180 seconds per epoch
    \item All the results are an average of $3$ runs.
\end{itemize}

After meticulous manual adjustment of hyperparameters, we have identified the ideal parameter configuration. In our exploration of batch sizes, ranging from $2$ to $8$, we settled on a batch size of $4$ due to computational limitations. We chose a learning rate of $5e-6$ with a weight decay of $1e-4$ as lower learning rates led to excessively slow training, while higher rates resulted in erratic learning behavior.

\end{document}